\documentclass[letterpaper, 10 pt, conference]{ieeeconf}  

\IEEEoverridecommandlockouts                              

\usepackage{multicol}
\usepackage[bookmarks=true]{hyperref}
\usepackage{color}
\usepackage{graphicx}
\usepackage{caption}
\usepackage{subcaption}
\usepackage{soul}
\usepackage{amsmath}
\usepackage{gensymb}
\usepackage{makecell}

\newcommand{\Fig}[1]{Fig. \ref{#1}}
\newcommand{\Eq}[1]{Eq. (\ref{#1})}
\newcommand{\Sec}[1]{Sec. \ref{#1}}

\title{\LARGE \bf
Active Image-based Modeling with a Toy Drone
}

\author{Rui Huang$^{1,3}$, Danping Zou$^{2}$, Richard Vaughan$^{1}$, Ping Tan$^{1}$
\thanks{$^{1}$School of Computing Science, Simon Fraser University, Canada.}%
\thanks{$^{2}$Shanghai Key Lab of Navigation and Location-based Services, Shanghai Jiao Tong University, China. This work is partially supported by NSFC Grant (No. 61402283).}
\thanks{$^{3}$The author is now with Alibaba AI Labs, China. Email: {\tt\small huangrui815@gmail.com}.}
}

\begin{document}

\maketitle
\thispagestyle{empty}
\pagestyle{empty}
\begin{abstract}
Image-based modeling techniques \cite{fuhrmann2014mve,altizure,pix4d} can now generate photo-realistic 3D models from images.
But it is up to users to provide high quality images with good coverage and view overlap,
which makes the data capturing process tedious and time consuming.
We seek to automate data capturing for image-based modeling.
The core of our system is an iterative linear method to solve the multi-view stereo (MVS) problem quickly and plan the Next-Best-View (NBV) effectively.
Our fast MVS algorithm 
enables online model reconstruction and quality assessment to determine the NBVs on the fly.
We test our system with a toy unmanned aerial vehicle (UAV) in simulated, indoor and outdoor experiments. Results show that our system improves the efficiency of data acquisition and ensures the completeness of the final model.
\end{abstract} 
\section{Introduction}\label{sec:introduction}

Image-based modeling methods create 3D models from digital images.
They often follow a standard pipeline of structure-from-motion (SfM)~\cite{snavely2006photo,agarwal2009building,wu2013towards}, multi-view stereo  (MVS)~\cite{goesele2007multi}, and surface modeling and texturing~\cite{fuhrmann2014floating,kazhdan2013screened}.
This pipeline has been extensively studied~\cite{tan2007image,xiao2009image,Muller2007},
and been demonstrated at different scales including desktop objects, buildings, and cities~\cite{goesele2007multi,jancosek2011multi}.
Now, both open source~\cite{fuhrmann2014mve} and commercial softwares~\cite{altizure,pix4d} are available to reconstruct high quality 3D model from images.

The results' quality of those systems strongly relies on the input images.
Under unfavorable conditions such as occlusion, motion blur, and poor illumination,
the user has to re-capture additional images to cover the missing part of the 3D model after the MVS step.
Unfortunately, existing MVS algorithms often take hours to reconstruct the hundreds of input images,
which makes the iteration of data capturing and modeling unbearably slow.
It is also difficult even for experienced users to determine the camera views to capture additional images,
which should remedy the missing parts and
keep sufficient view overlaps with existing images to let SfM and MVS work properly.


We present an active image-based modeling system with a toy unmanned aerial vehicle (UAV).
Our system puts image capturing in the optimization loop to actively plan the UAV's flight.
Specifically,
the user first sets a simple initial flight path. 
The captured images are processed by our fast MVS algorithm to estimate the rough shape of the object,
so that NBVs can be planned to let the UAV  take more images on the fly. 
Our system automatically iterates this process of image capturing, MVS reconstruction, and NBV planning 
until satisfactory accuracy has been achieved without the user's involvement.

We make two key technical contributions.
Firstly, we present a fast MVS algorithm that 
can reconstruct 100 images in about 20 seconds whereas conventional methods need hours on the same inputs.
This rapid MVS algorithm enables on-the-fly active image capture.
Secondly, 
we present a novel NBV planning method tailored for our MVS algorithm.
This method searches NBVs plane by plane, from high to low altitude, which helps to avoid collision as the UAV always flies above uncertain regions. Our complete system has been tested with a toy UAV in simulated, indoor, and outdoor scenes. Experimental results show that
our system collects sufficient images in less than half a hour to reconstruct a complete 3D model of a building-scale scene, 
which usually costs several hours in conventional MVS pipelines.

\section{Related Work}

\begin{figure*}
\centering
\includegraphics[width=0.9\textwidth]{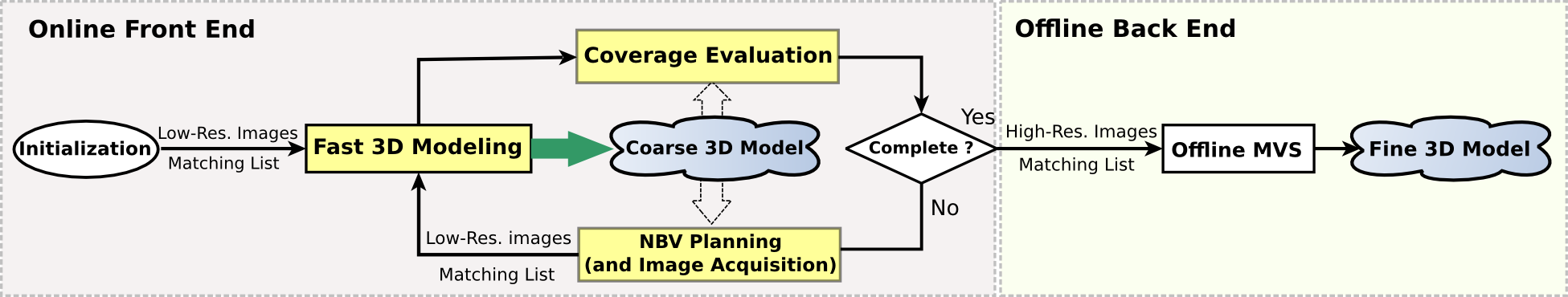}
\caption{The pipeline of our system, see \Sec{sec:overview} for more details.
}
\label{fig:pipeline}
\vspace{-0.2in}
\end{figure*}

\subsection{Active 3D Modeling System}
Active 3D modeling systems put data acquisition in the optimization loop,
using partial results to guide further data acquisition.
This problem has been studied with depth sensors~\cite{wu2014quality,kriegel2015efficient,khalfaoui2013efficient} on small scale objects.
Generalizing this idea to color cameras and to outdoor architectures is much harder.
Hoppe et al.~\cite{hoppe2012photogrammetric} have demonstrated such a system with a conventional MVS algorithm~\cite{furukawa2010towards},
where the NBVs are selected by the covariance of the mesh vertices.
Due to the poor efficiency of MVS, their system is not demonstrated for online processing.
In comparison, Mostegel et al.~\cite{mostegel2016uav} use machine learning techniques to predict the MVS quality without actually executing it.
In this way, they can also infer NBVs to improve the final MVS result.
Roberts  et al.~\cite{roberts2017submodular} present an image-modeling system that automatically plans flight to collect more images to improve the MVS reconstruction.
A similar system is also proposed in \cite{hepp2017plan3d}.
Both systems confront the same problem that their MVS components are too slow to achieve on-the-fly planning.
In comparison, our system is much more efficient, due to our fast MVS and  NBV algorithms.
It can finish active planning and image capturing in 20-30 minutes for a typical outdoor architecture.
To the best of our knowledge, our system is the first one to achieve this goal.

\subsection{Multiple-View Stereo (MVS)}
MVS aims to compute a per-pixel 3D point for each input image.
Classic MVS algorithms are often based on volumetric graph-cut~\cite{vogiatzis2007multiview,goesele2006multi},
level-set optimization~\cite{faugeras1998complete}, or iteratively matching-propagation~\cite{furukawa2010towards,lhuillier2005quasi}.
Due to the heavy optimization task,
MVS is the most time consuming process in the image-based  modeling pipeline,
and usually takes hours to reconstruct a scene at the scale of our examples.
The high computational complexity prevents its usage in online processing.
It is recently shown that dense piecewise planar surfaces can be  reconstructed from a single image and sparse SfM points~\cite{bodis2015superpixel}.
This method is fast as it only involves solving a linear equation. Inspired by this work, our method  is not limited to a single image and enforces multi-view consistency to further improve reconstruction quality.
This method can generate reconstruction with sufficient quality to guide the UAV to find NBVs with little delay.


\subsection{Next-Best-View (NBV) Planning}
Identifying an optimal viewpoint to produce a good 3D model is a classic problem in robotics~\cite{connolly1985determination,chen2011active}.
It is often difficult to precisely determine the NBV, because the actual 3D shape of the scene is unknown.
Many methods~\cite{kriegel2015efficient,khalfaoui2013efficient,dellepiane2013assisted} are heuristic,
relying on holes, open boundaries, or point densities to find NBVs.
Some approaches work well in relatively simple scenes,
but cannot deal with complex outdoor urban scenes, because those heuristics cannot generalize.
A recent method~\cite{wu2014quality} first does a Poisson surface reconstruction to estimate the 3D shape,
and then searches for NBVs accordingly.
It has been demonstrated with a laser scanner for  small desktop objects.
Our system also finds NBVs based on the Poisson surface, but under the more complicated MVS setting and in large outdoor scenes.

Another difficulty of the NBV problem is the large search space.
Typically, the solution space is uniformly quantized, e.g. into a 3D voxel grid~\cite{wu2014quality,kriegel2015efficient},
and each candidate camera pose needs to be evaluated to identify the NBVs.
It involves a huge amount of computation and is impractical for an online system.
Some methods~\cite{dunn2009next,trummer2010online} restrict the solution space to a sphere surface to reduce computation
at the cost of reducing the chance of finding the true NBVs.
Other methods~\cite{hoppe2012online} only consider candidate viewpoints with a constant distance to the object to deal with large scale objects.
In comparison, our system searches NBVs plane by plane, from high to low altitude, to speedup the process and to facilitate collision avoidance.

\section{System Overview}\label{sec:overview}

\begin{figure*}
\centering
\includegraphics[width=0.9\textwidth]{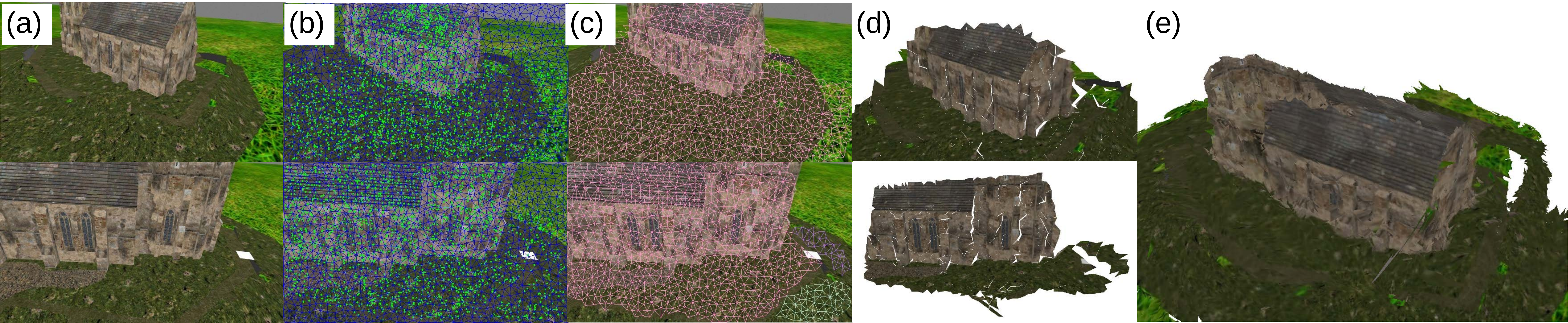}
\caption{Our novel linear piecewise planar MVS, see \Sec{sec:linear_MVS} for more details.
}
\vspace{-0.2in}
\label{fig:multiview_mesh}
\end{figure*}

Our system consists of an online front end and an offline back end as shown in \Fig{fig:pipeline}.
The front end controls the image capturing process to ensure good data coverage.
The back end takes an existing method~\cite{jancosek2011multi} to build a high quality 3D model from the captured images.
We focus on the front end that consists of mainly three components: \emph{Fast 3D Modeling}, \emph{Coverage Evaluation}, and \emph{NBV Planning} (and image acquisition).
The \emph{Fast 3D Modeling} component uses down-sampled images to quickly generate a coarse 3D model.
The \emph{Coverage Evaluation} component identifies places on the 3D model that require additional images.
When more images are needed, the \emph{NBV Planning} component determines a sparse set of camera views that are of the best chance to improve the 3D model and ensure view overlap. A flight path is then planned to drive the UAV to those positions.

\section{Fast 3D Modeling}
The \emph{Fast 3D Modeling} is called in the loop of the online process whenever more images are captured.
Its efficiency is therefore  critical to make the front end fast.
We propose a novel method to solve a dense reconstruction efficiently.
This method produces results with sufficient quality for the following \emph{Coverage Evaluation}.
\vspace{-1mm}

\subsection{Sparse Reconstruction}
We take a standard incremental SfM method~\cite{wu2011visualsfm} to calibrate all cameras and reconstruct a sparse set of 3D scene points.
For better efficiency, we only match nearby images whose GPS positions are within 2 meters.
When additional images are captured, the model is updated incrementally rather than recomputed from scratch.
\vspace{-1mm}

\subsection{Linear Dense MVS}\label{sec:linear_MVS}

Traditional MVS algorithms solve a per-pixel depth for each input image.
To speed up this process, we regularize the depth map to a piecewise linear surface.
This idea is exemplified in \Fig{fig:multiview_mesh}.
Specifically, as shown in \Fig{fig:multiview_mesh} (a-b), we divide each input image into polygons by an over-segmentation algorithm~\cite{duan2015image}.
Each polygon is  split into triangles via the Delaunay triangulation~\cite{fortune1992voronoi}.
These triangles are further clustered into connected regions covering SfM points as in \Fig{fig:multiview_mesh} (c).
Instead of solving a per-pixel depth for each image, we solve a per-vertex depth for each triangle,
which is formulated as a linear equation and fast to solve.
This produces a 3D triangle mesh for each input image as shown in \Fig{fig:multiview_mesh} (d).
We further require the triangle meshes from different views to agree with each other by enforcing the multi-view constraint.
Therefore, the results from different views can be naturally fused as in \Fig{fig:multiview_mesh} (e).
\subsubsection{MVS Formulation}
Consider a triangle $\{v_1, v_2 , v_3\}$. 
Following \cite{bodis2015superpixel}, we seek to compute the depth at $v_1, v_2$ and $v_3$.
Suppose $p$ is a point reconstructed by SfM, and $p$ is projected in the triangle $\{v_1, v_2 , v_3\}$.
The depth of $p$ can be interpolated by: $ d_p = \alpha_1 d_1 + \alpha_2 d_2 + \alpha_3 d_3$.
Here, $d_p$ and $d_k$ are the inverse depths of $p$ and $v_k, 1\leq k \leq 3$ respectively.
The weights $(\alpha_1, \alpha_2, \alpha_3)$ are the barycentric coordinates of $p$ in the triangle $\{v_1, v_2 , v_3\}$.
We minimize the energy
\begin{equation}
E_{sfm}(i)= (d_p - \alpha_1 d_1 - \alpha_2 d_2 - \alpha_3 d_3)^2
\label{eq:sfm}
\end{equation}
to enforce consistency with SfM points.
Here, $i$ is the index for the input image.

\begin{figure}
\centering
\includegraphics[width=0.50\textwidth]{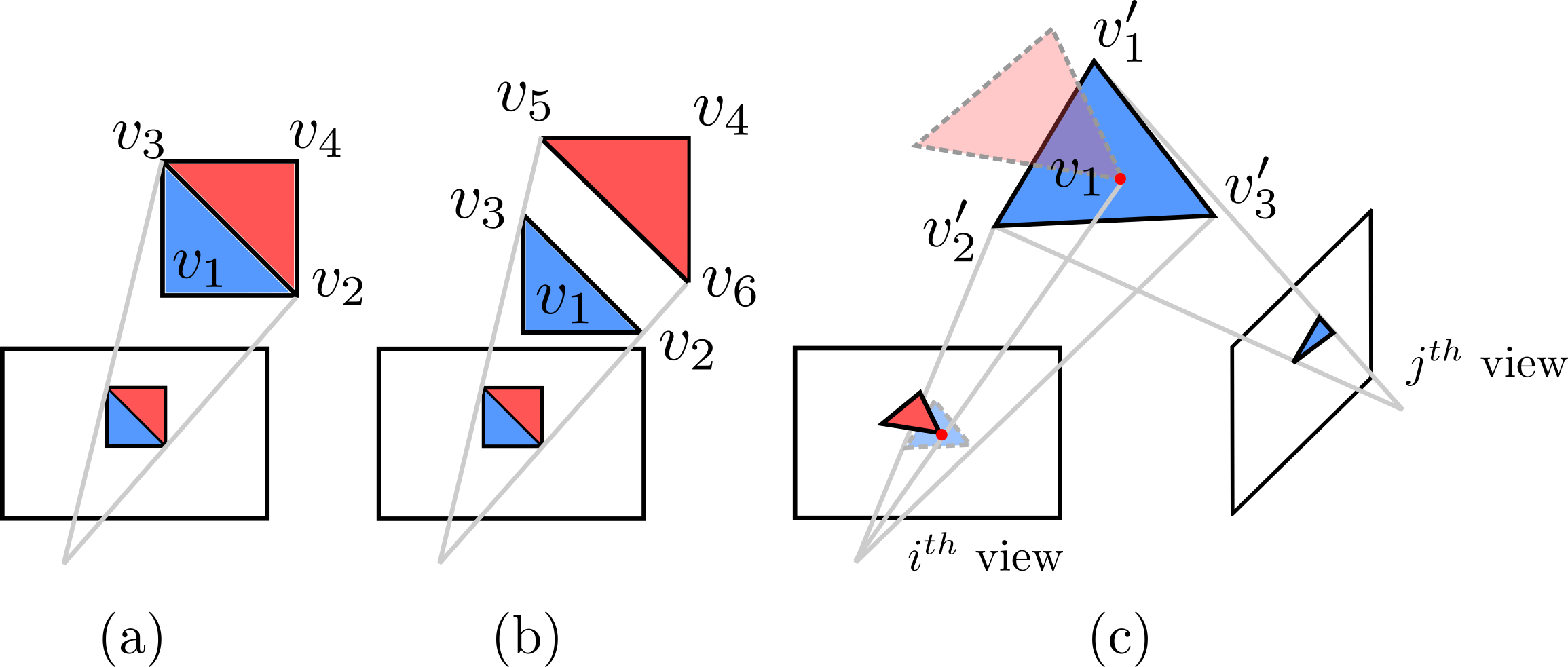}
\caption{Neighboring triangles are forced to have $C^0$ continuity with the parameterization in
 (a), while our parameterization in (b) (splitting $v_2, v_3$ in (a) to two vertices)  allows discontinuity.
 Triangles are projected to neighboring views to enforce multi-view consistency as shown in (c).
 }
 \vspace{-0.2in}
\label{fig:surface_representation}
\end{figure}

Now, consider two adjacent triangles, instead of using the parameterization in \cite{bodis2015superpixel} as shown in \Fig{fig:surface_representation} (a), 
we use the one shown in in \Fig{fig:surface_representation} (b).
Here, although $v_3$ and $v_5$ ($v_2$ and $v_6$) overlap in the image,
we parameterize them with two distinctive depths.
Our novel parameterization is critical to preserve depth discontinuity, 
because the edge $v_2v_3$ might be an occluding edge and the two triangles are at different depths.
We still favor local continuity 
by minimizing
\begin{equation}
E_{continuity}(i) = w_c ((d_3 - d_5)^2 + (d_2 - d_6)^2)
\label{eq:continuity}
\end{equation}
Here, the weight $w_c$ controls the smoothness strength, which is determined by the color difference of the triangles.

Meanwhile, we adopt the coplanar constraint from \cite{bodis2015superpixel},
\begin{equation}
E_{smooth}(i) = w_c (d_4 - \beta_1 d_1 - \beta_2 d_2 - \beta_3 d_3 )^2
\label{eq:smooth}
\end{equation}
Here, $d_4 =\beta_1 d_1 + \beta_2 d_2 + \beta_3 d_3$, and
$\{\beta_1, \beta_2 , \beta_3\}$ are the barycentric coordinates of $v_4$ with respect to the  triangle $\{v_1, v_2, v_3\}$.

We further introduce a novel multi-view consistency term.
Consider two neighboring views $i, j$ as shown in \Fig{fig:surface_representation} (c).
Suppose the triangle $\{v'_1, v'_2, v'_3\}$ in the $j$-th view is projected to cover the vertex $v_1$ in the $i$-th view.
We denote their inverse depth as $d'_1, d'_2$ and $d'_3$ respectively.
Then the depth at $v_1$ should be consistent with the one interpolated from $d'_1, d'_2$ and $d'_3$ according to the barycentric coordinates.
In other words, we should minimize the following energy,
\begin{equation}
E_{fusion}(i,j) = (d_1 - \gamma_1 d'_1 - \gamma_2 d'_2 - \gamma_3 d'_3)^2
\label{eq:fusion_term}
\end{equation}
where $(\gamma_1,\gamma_2,\gamma_3)$ are the barycentric coordinates of the $v_1$ in the projected triangle $\{v'_1, v'_2, v'_3\}$.
We typically consider a fixed number of neighboring views (e.g. 6 neighbors in our experiments) to build this constraint.

Putting all together, we can write the energy function as
\begin{equation}
\begin{split}
E =  \sum_i E_{sfm}(i) + \sum_i E_{continuity}(i)
\\+ \sum_i E_{smooth}(i) + \sum_{|i-j|<3} E_{fusion}(i,j)
\label{eq:surface_reconstruction}
\end{split}
\end{equation}
Note that except the fusion term, all other terms are linear functions of the vertex depths. 

\subsubsection{Optimization}\label{sec:optimization_details}
To solve \Eq{eq:surface_reconstruction},
we first solve a depth map at each view by ignoring the fusion term.
We then optimize the full energy to improve the individual depth maps for multi-view mesh fusion.

\noindent\textbf{Initialization} \hspace{5mm} This step simply discards the fusion term $E_{fusion}$ and minimizes the remaining terms per each view.
In this way, the original energy function becomes linear,
\begin{equation}
E = \|\mathbf{d}_{sfm} - \mathbf{A} \mathbf{d}\|^2 + \lambda_s \|\mathbf{B}\mathbf{d}\|^2_\mathbf{W_s} + \lambda_c \|\mathbf{C}\mathbf{d}\|^2_\mathbf{W_c}
\label{eq:surface_reconstruction_init}
\end{equation}

\noindent Here $\mathbf{d}_{sfm}$ is a $N_s\times1$ vector representing the inverse depths of all SfM points and $\mathbf{d}$ is a $N_v \times 1$ vector of unknown inverse depth of all mesh vertices. $\mathbf{A}$ is a $N_s\times N_v$ sparse matrix where each row contains the barycentric coordinates as in \Eq{eq:sfm}. $\mathbf{B}$ is a sparse matrix collecting all the smoothness constraints as described in \Eq{eq:smooth}. Each row of $\mathbf{C}$ consists of only $1$ and $-1$ to describe the continuity constraint defined in \Eq{eq:continuity}. Both $\mathbf{W}_s$ and $\mathbf{W}_c$ are diagonal matrices consisting of color difference penalty $w_c$ between adjacent triangles. $\lambda_s$ and $\lambda_c$ are weights of each constraint. \Eq{eq:surface_reconstruction_init} can be efficiently minimized by a linear solver.



\noindent\textbf{Confidence Evaluation} \hspace{5mm} The initialization step can generate large errors, especially at occlusion edges.
We fuse results from other views to help reduce errors.
To facilitate fusion, we compute a confidence score at each triangle to measure its quality derived from three cues: position consistency, normal consistency, and front parallelism.

\begin{figure}
\centering
\includegraphics[width=0.4\textwidth]{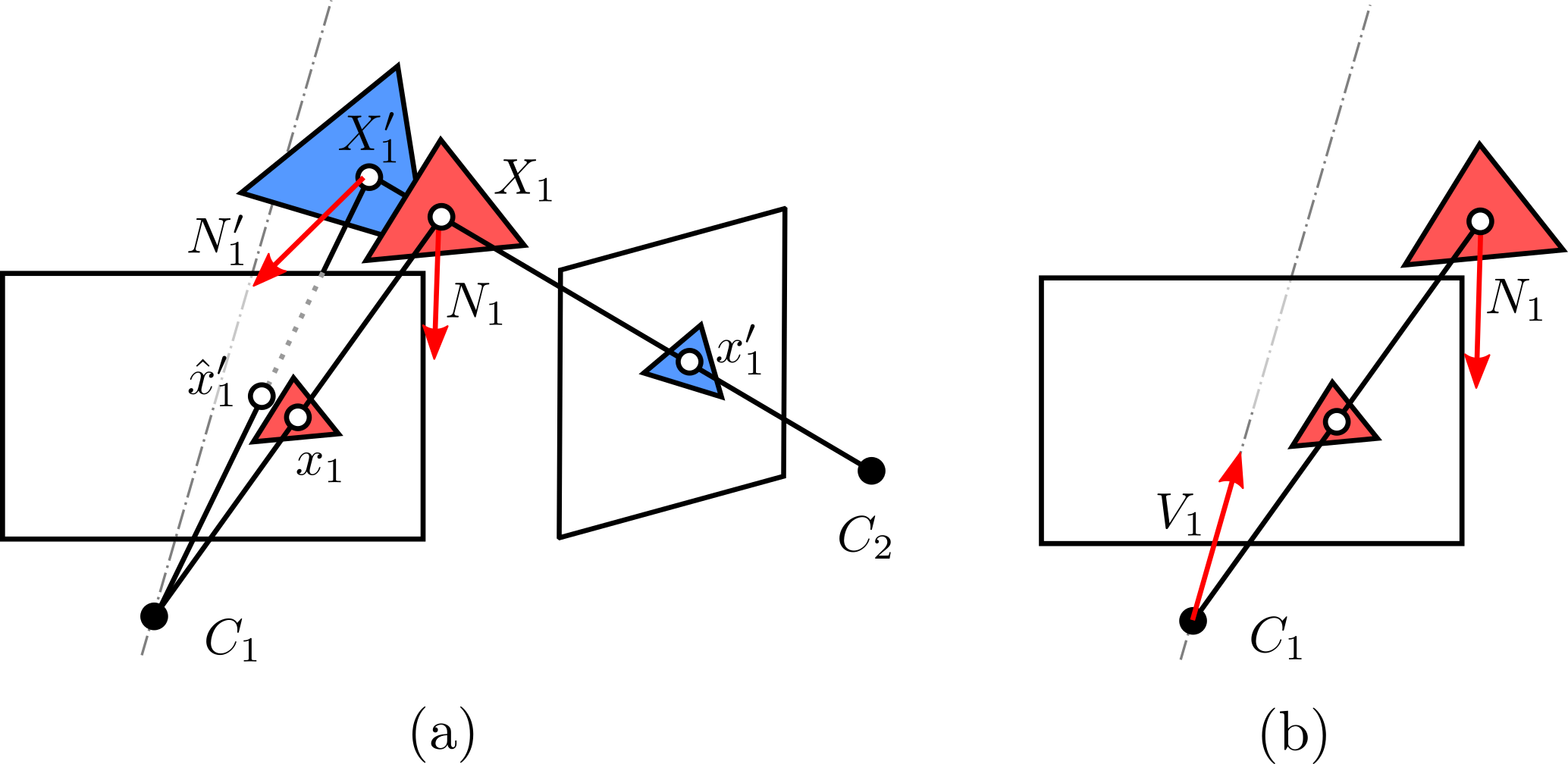}
\caption{Confidence evaluation based on cues from: (a). Position/normal consistency; (b). Front parallelism. }
\label{fig:confidence_evaluation}
\end{figure}

\emph{Position Consistency}:  As shown in \Fig{fig:confidence_evaluation} (a), suppose a triangle is reconstructed from the view $C_1$ and its centroid $X_1$ is projected at $x_1$.
To ensure the consistency, we project $X_1$ to a neighboring view $C_2$ at $x'_1$ and find the corresponding 3D position $X'_1$ from the depth map of $C_2$.
Let $\hat{x}'_1$ be the projection of $X'_1$ on the ${C_1}'s$ image plane.
The smaller distance between $x_1$ and $\hat{x}'_1$, i.e. $e_p=\|x_1 - \hat{x}'_1\|$, indicates better consistency.

\emph{Normal Consistency}: As shown in \Fig{fig:confidence_evaluation} (a), let $N_1$ be the normal direction of the triangle reconstructed from $C_1$ and $N'_1$ be the normal of corresponding triangle from $C_2$.
We check the normal consistency by measuring the difference between $N_1$ and $N'_1$, i.e. $e_n = \arccos(N_1^TN'_1).$

\emph{Front Parallelism}: Generally speaking, when the viewing direction of a camera is perpendicular to the object surface,
SfM algorithms tend to produce more reliable 3D points.
From this observation, we additionally measure the angle between the viewing direction $V_1$ of the camera and the estimated face normal $N_1$ as shown in \Fig{fig:confidence_evaluation} (b),
i.e. $e_v = \arccos(N_1^T V_1)$ to evaluate the reconstruction quality.

\begin{figure}
\centering
\includegraphics[width=0.4\textwidth]{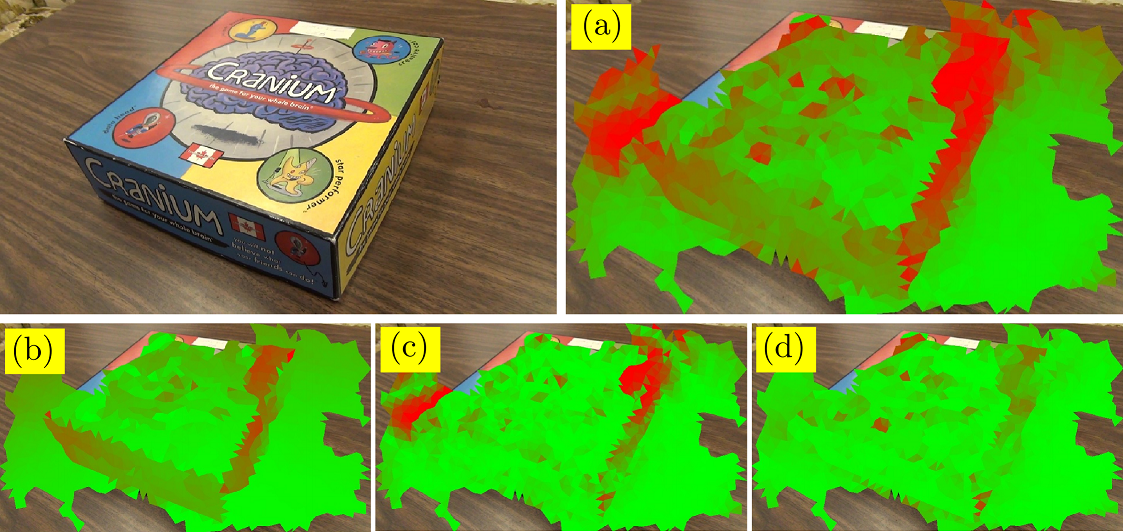}
\caption{An input image and (a) its depth confidence map, which is computed according to: (b) position consistency, (c) surface normal consistency, and (d) front parallelism.
High/low confidence regions are shown in green/red.}
\vspace{-0.2in}
\label{fig:confidence_map}
\end{figure}

For each triangle, we evaluate the above confidence measurements with respect to $N_{adj}$ neighboring views.
We take the mean $\bar{e}_p$ and $\bar{e}_n$ of $N_{adj}$ views and define the overall confidence measurement as
\begin{equation}
\Gamma = \exp(-\bar{e}_p/\sigma_p)\exp(-\bar{e}_n/\sigma_n)(1-\exp(-\cos^2(e_v)/\sigma_v))
\label{eq:confidence_score}
\end{equation}
where constants $\sigma_p,\sigma_n$ and $\sigma_v$ control the weight of each confidence measurement.
\Fig{fig:confidence_map} illustrates the confidence map of a surface.

\noindent\textbf{Multi-view Depth Fusion} \hspace{5mm} Once a confidence map is computed for each view, we start to fuse the depth maps by solving \Eq{eq:surface_reconstruction}.
We still solve the depth map one view at a time.
Consider a triangular face $f$  reconstructed from view $i$.
If its confidence score is higher than $0.5$, we leave its vertex fixed in \Eq{eq:surface_reconstruction_init}.
Otherwise, we optimize its depth by registering it to the corresponding triangles in other views by minimizing the fusion term.
The corresponding triangles are searched based on the space, normal and color proximity. We can optimize the fusion term together with \Eq{eq:surface_reconstruction_init} linearly,
\begin{equation}
\|\mathbf{d}_{sfm} - \mathbf{A} \mathbf{d}\|^2 + \lambda_s \|\mathbf{B}\mathbf{d}\|^2_\mathbf{W_s} + \lambda_c \|\mathbf{C}\mathbf{d}\|^2_\mathbf{W_c} \\+ \lambda_u \| \mathbf{d}_{ref} - \mathbf{R}\mathbf{d}_L\|^2
\label{eq:surface_reconstruction_fusion}
\end{equation}
where $\mathbf{R}$ and $\mathbf{d}_{ref}$ are both derived from \Eq{eq:fusion_term}, which respectively represents an identity matrix and depths of the corresponding triangles. $\lambda_u$ is the weight of fusion term.

\begin{figure}
\centering
\includegraphics[width=0.45\textwidth]{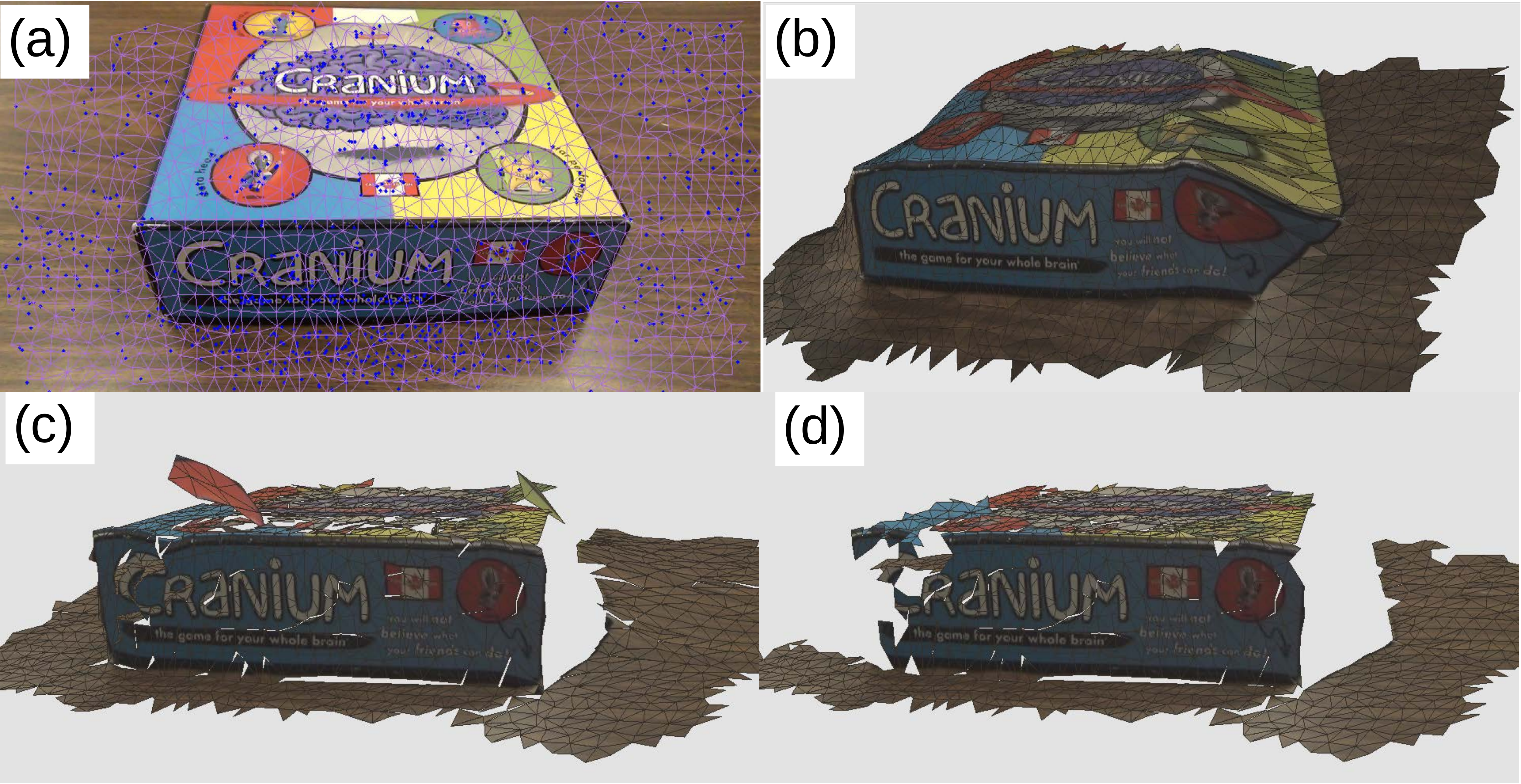}
\caption{For an input (a), (b) is the result from \cite{bodis2015superpixel},
(c) and (d) are our results without and with multi-view consistency.
}
\vspace{-0.2in}
\label{fig:single-view-reconstruction}
\end{figure}

We optimize \Eq{eq:surface_reconstruction_fusion} for each view to enforce mesh consistency cross neighboring views. Simply putting together all the meshes solved from single view, we are able to obtain a complete triangular mesh as shown in \Fig{fig:multiview_mesh} (e). \Fig{fig:single-view-reconstruction} compares our method with \cite{bodis2015superpixel}.
For the input image in (a), (b) shows the result by the method in \cite{bodis2015superpixel}. 
Due to its parameterization, the discontinuity between the box and the desktop cannot be preserved. 
\Fig{fig:single-view-reconstruction} (c) demonstrates the result without multi-view consistency.
Notice that depth discontinuity is preserved, but some of the triangles are still incorrect, due to insufficient SfM points in those regions.
Lastly, \Fig{fig:single-view-reconstruction} (d) shows our final result where noisy triangles are corrected.

\section{Active Image Acquisition}
After multi-view depth fusion, we obtain a fused triangular mesh. In the following, a coverage evaluation method is proposed to identify poorly reconstructed regions. 
Then we introduce our Next-Best-View (NBV) algorithm to capture additional viewpoints which improve the reconstructed model. Lastly, a path planning approach is proposed to connect the NBVs. The UAV will follow the planned path to capture additional images at those NBVs.

\subsection{Coverage Evaluation}
\label{sec:coverage_evaluation}
\begin{figure}
\includegraphics[width=0.48\textwidth]{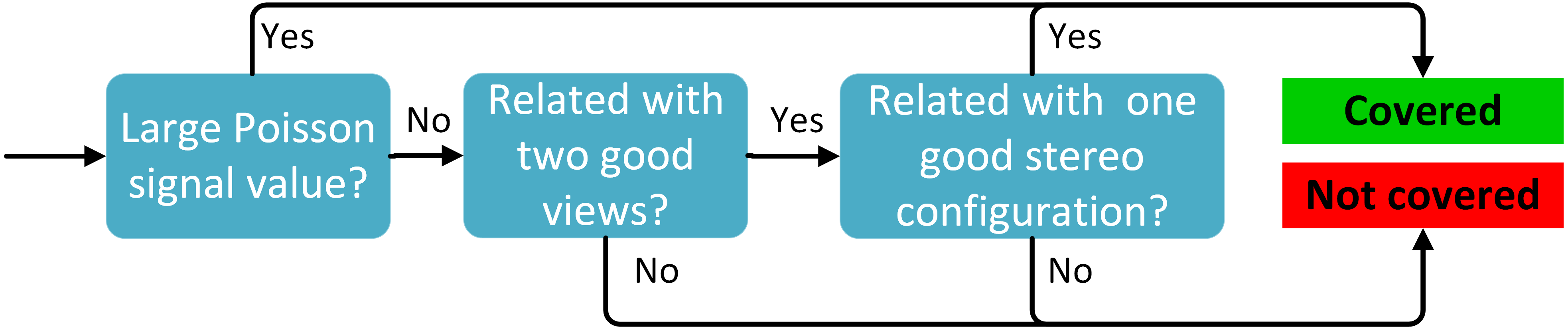}
\caption{The pipeline of classifying covered iso-points.}
\label{fig:coverage_classification}
\vspace{-0.2in}
\end{figure}

Coverage evaluation aims to identify under-sampled regions to facilitate the NBV selection. 
Our approach is inspired by the method in~\cite{wu2014quality} which is designed for laser-scanners.
It applies Poisson surface reconstruction\cite{kazhdan2013screened} to the point clouds and uses the Poisson signal value and smoothness to identify regions with poor coverage.
We adapt this idea to work with points reconstructed by our MVS method.
Intuitively, a surface is better reconstructed when it has higher resolution in the image.
Also, to ensure a good 3D reconstruction, a surface should be observed in at least two views with a reasonable view-span.
Based on these two observations, we uniformly sample the reconstructed surface by Poisson disk sampling~\cite{bridson2007fast} to get sampled points named as \emph{iso-points}. We then classify each iso-point as `covered' or `un-covered' according to the pipeline shown in \Fig{fig:coverage_classification}.
Basically, large Poisson signal value means the point is well reconstructed, i.e. `covered'.
We further evaluate iso-points with low signal values.
Let $A_p$ be the area of a small disk at an \emph{iso-point} $p$ in the 3D space and $a_p(v)$ be the area of its projection on the image.
We define the \emph{projection ratio} of this iso-point as $\gamma_p(v) = a_p(v)/A_p$.
A large \emph{projection ratio} means a better chance of high-quality reconstruction.
For each iso-point, we mark the views with a large projection ratio ($> \gamma_{min}$)  as a `good' view.
If an iso-point is seen by less than two `good' views, it is marked as `uncovered'.
Lastly, if the angle between the `good' views is within a proper range $[\theta_{min} = 2\degree, \theta_{max} = 30\degree]$,
these `good' views form a robust stereo configuration and the point is marked as `covered', otherwise 'uncovered'.

\begin{figure}
\centering
\includegraphics[width=0.48\textwidth]{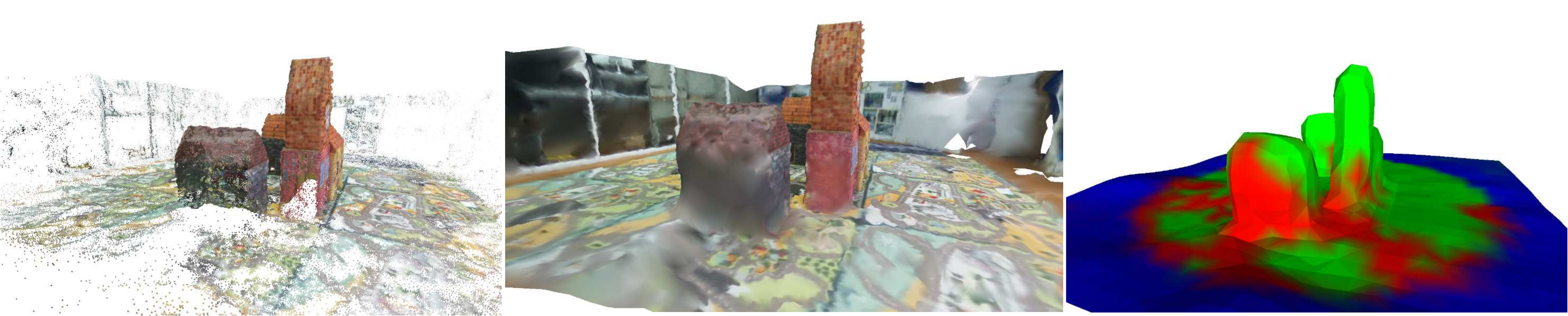}
\caption{An example of the coverage map. Left: reconstructed point cloud from our fast MVS;
middle: 3D model from the offline back end modeling method~\cite{jancosek2011multi};
right: a color coded coverage map covered (green), uncovered (red) and ignored (blue) areas.
}
\vspace{-0.2in}
\label{fig:coverage_score}
\end{figure}

\Fig{fig:coverage_score} shows a coverage map.
From left to right, those are the reconstructed triangle vertices by our fast MVS method,
a 3D model by the offline MVS method~\cite{jancosek2011multi} and the evaluated coverage map,
where `covered' and `uncovered' points are shown in green and red respectively.
We can see that the 3D model in the middle have poor quality at `uncovered' red regions shown in the right.

\subsection{NBV Planning and Flight Path Planning}
The NBV problem is NP-hard and is often solved approximately by greedy algorithms.
Though there are recent NBV algorithms for small desktop objects with a laser scanner~\cite{wu2014quality, kriegel2015efficient,khalfaoui2013efficient},
these methods are still too computational expensive for our large-scale outdoor scenes.
The computational complexity comes from the solution space quantization and candidate view evaluation. 




Our candidate view evaluation is efficient, thanks to our effective coverage evaluation in \Sec{sec:coverage_evaluation}.
A good NBV should improve the reconstruction at `uncovered' points.
Thus, we use the sum of \emph{projection ratio} of observed `uncovered' iso-points to evaluate a view candidate.

In principle, we need to discretize a 5D space (pitch, yaw and $x,y,z$ coordinates) to search for NBVs.
For more efficient search, we fix the altitude $z$ for each pitch angle.
In this way, we reduce the search space from 5D to 4D.
Specifically, we quantize 12 camera pitch angles uniformly between $[-30\degree, 30\degree]$.
For each sampled pitch angle $\theta_p$,
we determine a desired altitude such that the highest `uncovered' iso-point is projected to the image center according to $\theta_p$ and a predetermined safe distance. 
This altitude defines a plane $H_{\theta}$ the UAV should fly in, which is above all uncertain points and helps to avoid collisions.
We then evaluate NBVs in each plane $H_{\theta}$.
We uniformly sample a 2D grid of viewpoints in the plane and sample 8 yaw angles at each viewpoint.
On each plane $H_{\theta}$, we select $N_{nbv}$ ($=5$ in our experiments) local maximums as our NBVs with non-maximum suppression with radius of $1$ meter.
We further exclude candidates that are too far away from existing views to ensure successful feature matching in SfM.
By too far, we mean the pitch and yaw angles differ by more than $15\degree$ or the position differ by more than 0.5 meters in indoor (or 3 meters in outdoor) experiments.We experimentally determine all the parameters based on the camera on the UAV.
Finally, among the 12 sampled pitch angles, we choose the one (and its associated altitude) that can bring the NBV with highest score.

To avoid collision, we first discretize the plane of NBVs into regular cells. then mark those cells whose distance to the reconstructed 3D model is smaller than the safe distance as `occupied' and mark the other cells as `free'.
In the `free' cells, A-star algorithm~\cite{hart1968formal} is used to generate a path connecting the NBVs starting from UAV's current position. When arriving at an NBV, the UAV adjusts its camera angles to take an image before moving to the next nearest NBV.



\begin{figure*}
\centering
\includegraphics[width=0.95\textwidth]{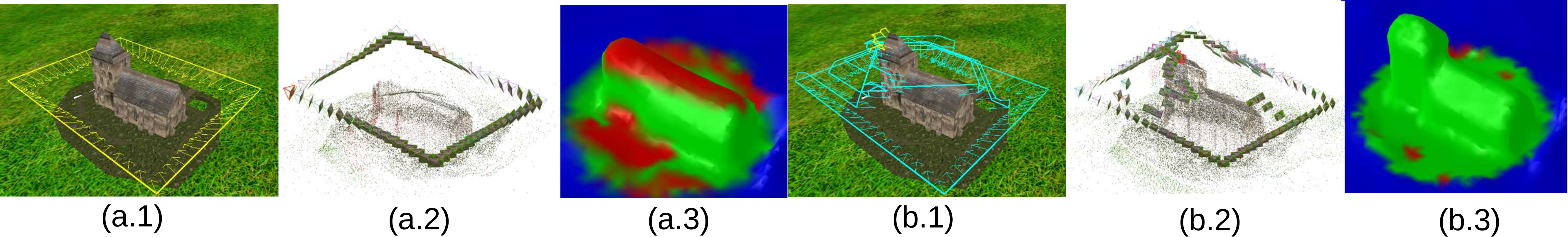}
\caption{The \emph{Church} example: flight paths (1), SfM results (2), and coverage maps (3) after the first (a) and final (b) iteration.}
\label{fig:church-path-mesh-score}
\vspace{-0.1in}
\end{figure*}

\begin{figure*}
	\centering
	\includegraphics[width=0.95\textwidth]{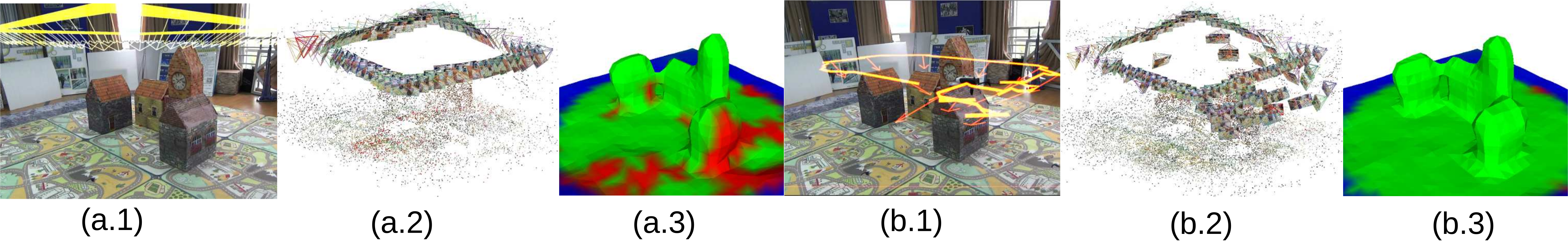}
	\caption{The \emph{Indoor} example with a Bebop drone in a Vicon room. }
	\label{fig:indoor18-path-mesh-score}
	\vspace{-0.2in}
\end{figure*}



\section{Experiments}
We verify our system with both simulated and real experiments.
For simulation, we use the Gazebo simulation platform \cite{gazebo} and 3D architecture models from 3D Warehouse \cite{3dWarehouse}.
For real experiments, we tested in both indoor Vicon rooms and outdoor open fields with a Bebop drone,
where the UAV's localization is achieved by Vicon and GPS respectively.
The drone sends low resolution ($640\times368$) images via a WiFi link to an Asus GL552 laptop which sends control commands back. High-resolution ($1920\times1080$) images are saved on board for offline processing.

\subsection{Simulated Experiments}
\subsubsection{Church}
\begin{figure}
\centering
\includegraphics[width=0.48\textwidth]{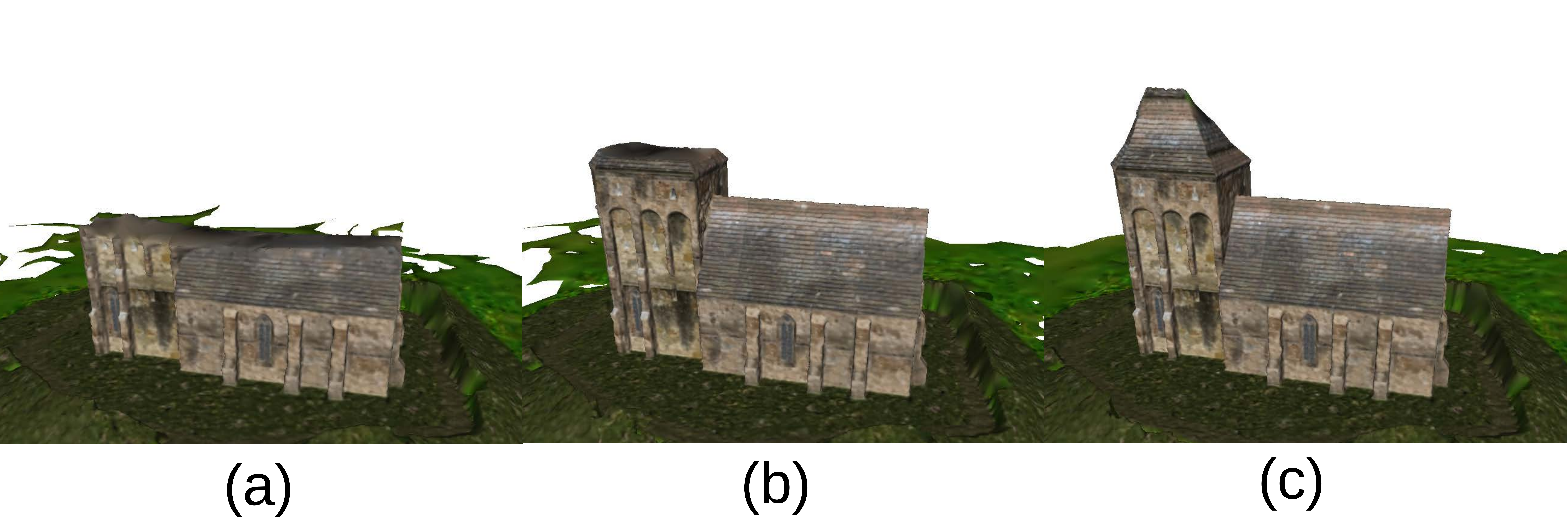}
\caption{The \emph{Church} example: High resolution models (a-c) produced offline from CMP-MVS after the 1st, 4th and final iteration of image capture.}
\label{fig:church-cmp-res}
\vspace{-0.2in}
\end{figure}

As shown in \Fig{fig:church-path-mesh-score} (a.1), the UAV takes images at some initial viewpoints sampled along a rectangular path around the object of interest.
The camera's initial pitch angle is $30\degree$ downward.
\Fig{fig:church-path-mesh-score} (a.2) shows the results from SfM with 64 low-res ($640\times368$) images at the initial positions,
and (a.3) shows the coverage evaluation result.
From these images,
a 3D model can be generated by the offline CMP-MVS method \cite{jancosek2011multi} as shown in \Fig{fig:church-cmp-res} (a).
It is clear that `uncovered' vertices correspond to poor final 3D modeling, e.g. the missing roof.
This \emph{Church} example is fully covered in 6 iterations of image capturing.
\Fig{fig:church-path-mesh-score} (b.1) shows the additional flight paths and sampled viewpoints.
The 3D positions and orientations of these views can be seen in (b.2).
The coverage result in (b.3) suggests the model is well covered by these images.
This can be verified in \Fig{fig:church-cmp-res}, which shows high-res models from CMP-MVS \cite{jancosek2011multi} after the 1st, 4th and final iteration.
The missing part is reconstructed gradually, and good result is finally achieved in \Fig{fig:church-cmp-res} (c).

\subsection{Indoor Experiments}
\begin{figure}[h!]
\centering
\includegraphics[width=0.48\textwidth]{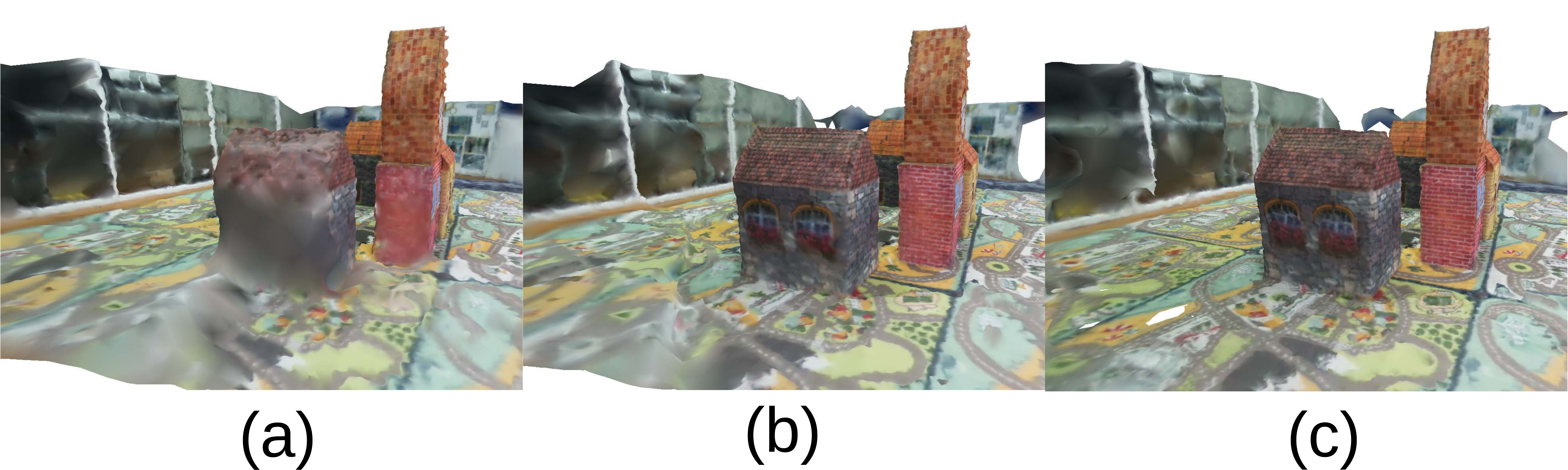}
\caption{The \emph{Indoor} example: CMP-MVS results (a-c) after iteration 1-3. }
\label{fig:indoor18-cmp-res}
\vspace{-0.15in}
\end{figure}

We further verify our system in a Vicon room decorated with some cardboard boxes resembling buildings. The UAV's pose is captured by Vicon for real-time control. The UAV transmits low-res ($640\times368$) images in real-time to the ground station while keeps high-res ($1920\times1080$) images on board for offline process.
65 images are captured from the initial scan.
This model is covered in 3 iterations.
\Fig{fig:indoor18-path-mesh-score} shows the result at the first and last iteration.
Again, (a.1) and (b.1) are the flight paths and view orientations, (a.2) and (b.2) are SfM results.
(a.3) and (b.3) are the coverage evaluation result.
The 3D models generated by CMP-MVS \cite{jancosek2011multi} are shown in \Fig{fig:indoor18-cmp-res}. We can tell the 3D model quality is well predicted by the coverage map in \Fig{fig:indoor18-path-mesh-score} (a.3).
Our NBVs successfully identify a small set of images to improve the model,
which is verified by both the coverage map in \Fig{fig:indoor18-path-mesh-score} (b.3) and the model improvement shown in \Fig{fig:indoor18-cmp-res} (c).

\begin{figure}[!ht]
	\centering
	\includegraphics[width=0.5\textwidth]{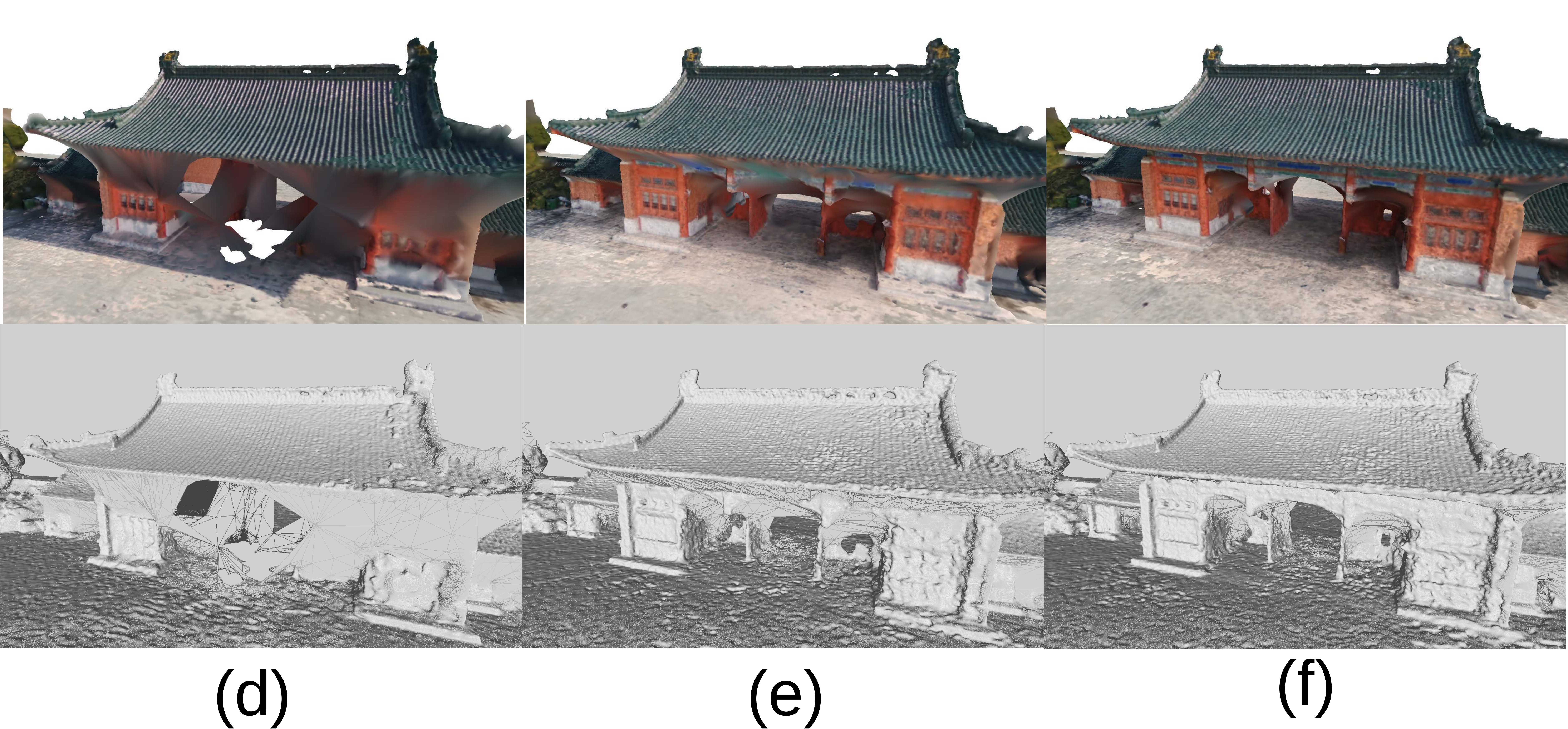}
	\caption{Offline MVS results of the \emph{Asian Building} example after the 1st, 3rd, and 5th iteration (from left to right).}
	\label{fig:eagle-CMP_MVS results}
	\vspace{-0.2in}
\end{figure}

\begin{figure*}
\centering
\includegraphics[width=0.95\textwidth]{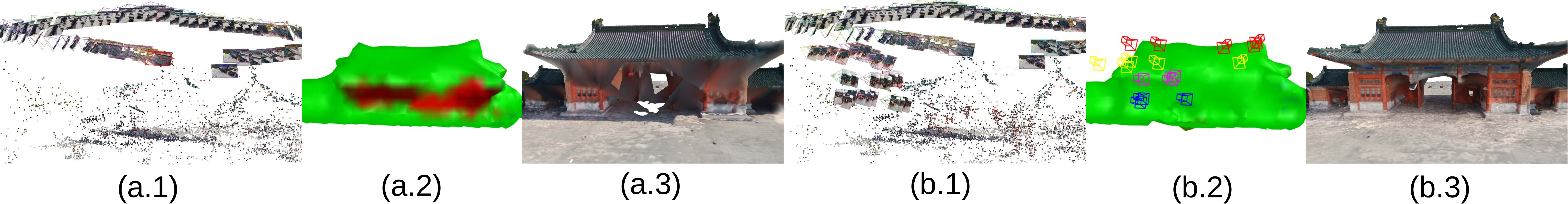}
\caption{
The \emph{Asian Building} example: Results after the 1st (a.1-3) and last (b.1-3) iteration.
}
\label{fig:eagle-nbvs}
\vspace{-0.2in}
\end{figure*}

\subsection{Outdoor Experiments}
\vspace{1mm}
We further demonstrate our system with real examples in outdoor open areas.
The Asian building in \Fig{fig:eagle-nbvs} is an entrance gate of $15$-meter height.
We initialize image capture with an enclosing rectangular flight path at $18$-meter height (see the reconstructed camera trajectory in \Fig{fig:eagle-nbvs} (a.1)).
Our system plans 4 more iterations of image capturing.
\Fig{fig:eagle-nbvs} (a.2) shows the color coded surface after coverage evaluation.
In the initialization, some parts under the roof (\Fig{fig:eagle-nbvs} (a.3)) are not covered by the input images,
as indicated in red. Our system successfully guides the UAV to gradually lower down and raise its camera pitch angle to capture those regions. This process can be seen from the reconstructed NBVs in \Fig{fig:eagle-nbvs} (b.1).
After additional images are taken, the coverage map turns to green in \Fig{fig:eagle-nbvs} (b.2).
\Fig{fig:eagle-nbvs} (b.3) shows the high quality 3D model produced by the offline modeling system with all images.
For a better validation, \Fig{fig:eagle-CMP_MVS results} (a-c) presents the 3D models generated by CMP-MVS from the high-resolution images after the 1st, 3rd and final iteration.

\subsection{Running Time}
We report our running time on the 1st iteration (81 images captured) of the outdoor example as a reference.
Our system takes 21.5 sec for the fast MVS and then calls an external Poisson surface reconstruction application~\cite{poisson-software} to generate a mesh using 24.8 sec. The NBV planning takes 16.4 sec and a path is planned in less than 0.01 sec.
In comparison, the MVE \cite{goesele2006multi} and CMP-MVS \cite{jancosek2011multi} systems take about 30 minutes and 40 minutes to reconstruct a mesh,
though at high quality. 
The recent GPU accelerated MVS system~\cite{galliani2015massively} takes 235.87 seconds. All systems are tested on the same laptop with i7-CPU, GTX850M-GPU and 16 GB RAM.


\subsection{Comparison of Coverage Evaluation}
\begin{figure}
\centering
\includegraphics[width=0.4\textwidth]{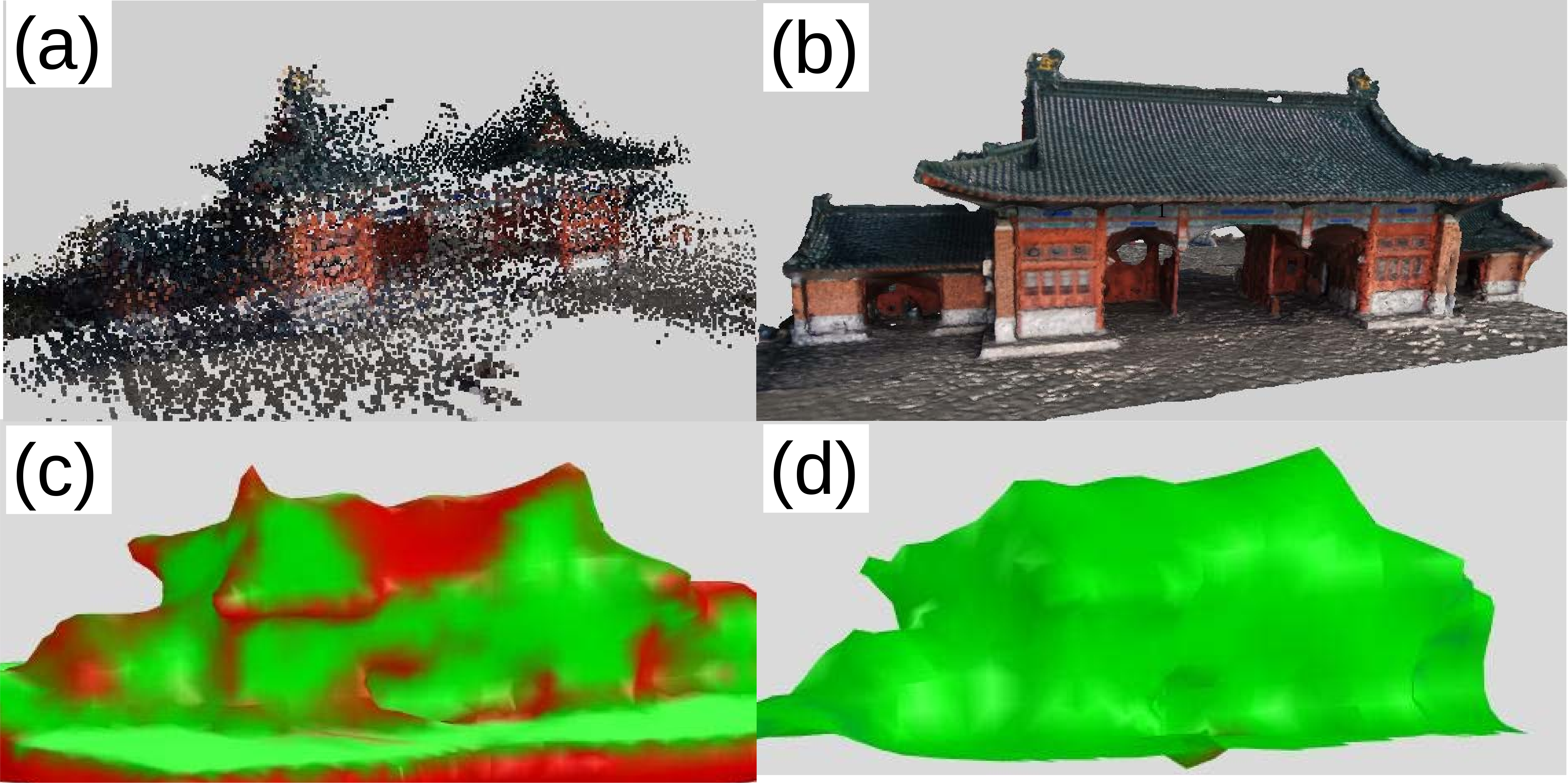}
\caption{Comparison of coverage evaluation.}
\label{fig:comparison_coverage}
\vspace{-0.1in}
\end{figure}

Given the MVS reconstructed points in \Fig{fig:comparison_coverage} (a), 
(c) and (d) are the coverage map computed by our method and the method in \cite{wu2014quality}, which is designed for laser scanners.
The CMP-MVS result in (b) clearly indicates the image coverage is sufficient to produce a high quality 3D model.
This is faithfully captured by our coverage map in (d).
However, the laser-based method \cite{wu2014quality} only considers the density and smoothness of the input point cloud.
Thus, it marks the roof as uncovered in red.


\subsection{Quantitative Evaluation}
In this section, we perform quantitative evaluation on the effectiveness and efficiency of our system.

\subsubsection{Manually-designed Flight Plans}
\begin{figure}
\centering
\includegraphics[width=0.48\textwidth]{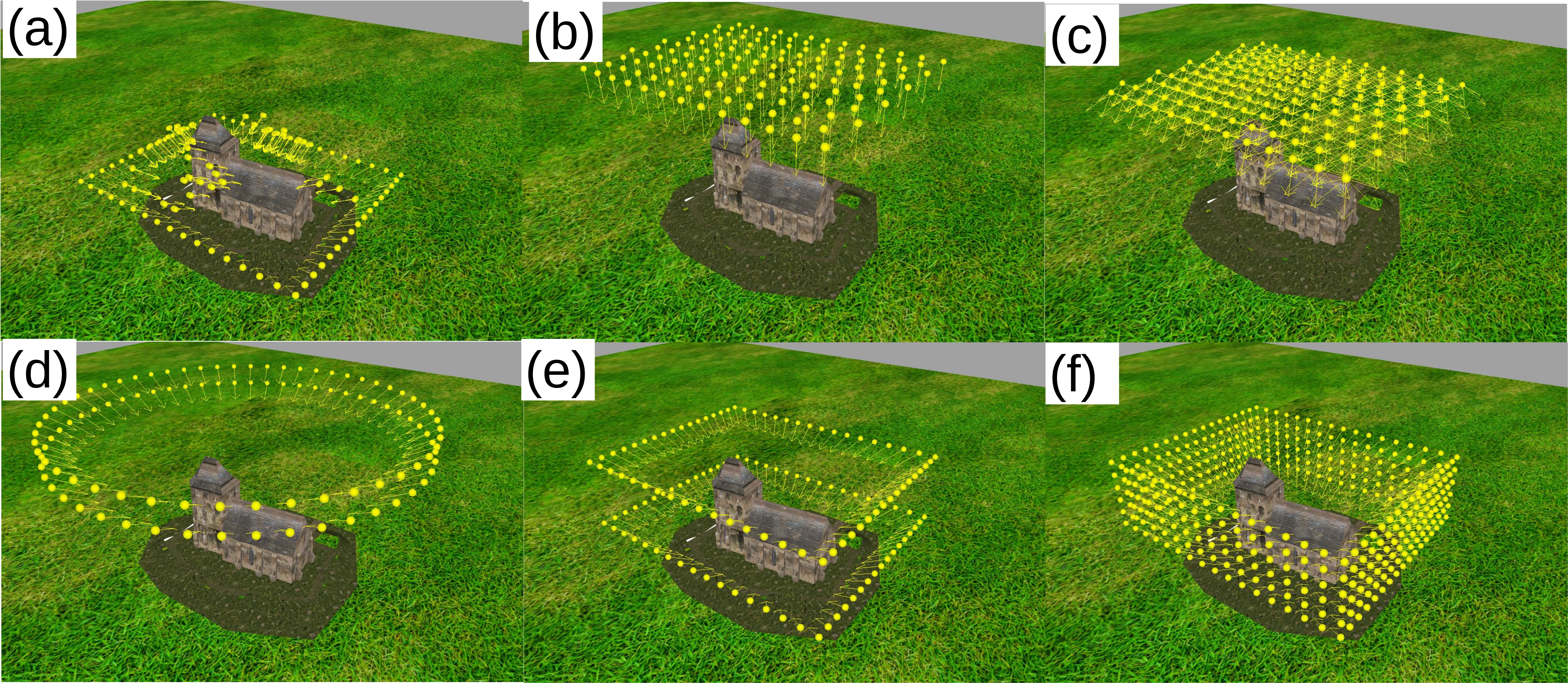}
\caption{The automated (a) and manual (b-f) capture plans. }
\label{fig:comparison_path}
\vspace{-0.2in}
\end{figure}

We compare with flight plans commonly used in aerial mapping. 
As shown in \Fig{fig:comparison_path}, we compare our method (a) with 5 other manually designed flight plans shown in (b--f).
For easy of reference, we name them as: \emph{Grid-Downward} (b), \emph{Grid-Multi-Angle} (c),
\emph{2-Circles} (d), \emph{Rectangle-Sparse} (e), \emph{Rectangle-Dense} (f).


\subsubsection{Model Alignment}
For each set of images obtained from the flight plans, we reconstruct the camera poses using \emph{Visual SfM}~\cite{wu2013towards} and generate 3D models using \emph{CMP-MVS}~\cite{jancosek2011multi}. A dense point cloud $P_\mathrm{M}$ is sampled on the generated 3D model using Poisson disk sampling~\cite{bridson2007fast}. Next, we align the reconstructed 3D model with the ground-truth model. Firstly, the reconstructed camera poses are registered to the ground-truth camera poses in the simulator or GPS coordinates in the outdoor scene, which generates an initial transformation $T_{init}$. Starting from $T_{init}$, we use an extension of ICP~\cite{umeyama1991least} to refine the registration. 

\subsubsection{Evaluation Metric}
We use two measures, \emph{model accuracy} and \emph{model completeness}, for the evaluation. The reconstructed model can be incomplete and the Poisson surface reconstruction interpolates the under-sampled regions. These regions should not be considered when we evaluate the \emph{model accuracy}. Therefore, we sample points $P_\mathrm{G}$ on the ground truth model $M_\mathrm{G}$. For each sample point, we search for the closest vertex $V_\mathrm{R}$ on the reconstructed model. The \emph{model accuracy} can be measured by the mean and RMS errors.

The \emph{completeness} is evaluated by the percentage of sample points with distance error smaller than a distance threshold $d$, which can be expressed as,
\begin{equation}
Completeness = \frac{100}{|P_\mathrm{G}|}\sum_{i}^{}\left[||P^i_\mathrm{G} - V^i_\mathrm{R}|| < d\right]
\label{eq:completeness}
\end{equation}
where the $[\cdot]$ is the Iverson bracket.

\begin{table}
\centering
\renewcommand{\arraystretch}{1.8}
    \begin{tabular}{ | c | c | c | c | c | c | c |}
    \hline
    Flight Plans  & \textbf{(a)}    & (b)         & (c)       & (d)     & (e)        & \textbf{(f)}        	\\ \hline
    \# of Views   & \textbf{108}    & 143         & 858       & 105     & 130        & \textbf{390}         \\ \hline
    Mean error (inch)    & \textbf{1.89}   & 3.55 	      & 3.05      & 3.65    & 2.24   	 & \textbf{1.79}		\\ \hline
    RMS error (inch)    & \textbf{2.93}	& 4.52 	      & 3.81 	  & 4.50    & 3.05  	 & \textbf{2.62}		\\ \hline
    1-inch completeness (\%)  & \textbf{35.8} & 4.7 	  & 5.3	    & 2.8        & 15.8	    & \textbf{28.6}  		\\ \hline
    2-inch completeness (\%) & \textbf{76.1} & 30.2	  & 31.9	& 20.3       & 64.2	    & \textbf{80.6}         \\ \hline
    \end{tabular}
\caption{Comparison of model accuracy and completeness between different plans.}
\vspace{-0.2in}
\label{table:comparison_accuracy}
\end{table}


\subsubsection{Evaluation Results}


\paragraph*{\emph{Simulated Experiments}} The model accuracy and completeness comparison is shown in Table~\ref{table:comparison_accuracy}. Our automated system, i.e. (a), captures 108 views in 5 iterations. The generated model accuracy is 1.89 inches of mean error and 2.93 inches of RMS error. \emph{Rectangle-Dense}, i.e. (f), is the only plan which produces better model accuracy (Mean error = 1.79, RMS error = 2.62) while 390 views need to be captured. 
The completeness results are in consistent with the model accuracy. \emph{Rectangle-Dense} (f) achieves better completeness at the cost of using many more images.

\begin{figure}
\centering
\includegraphics[width=1.0\linewidth]{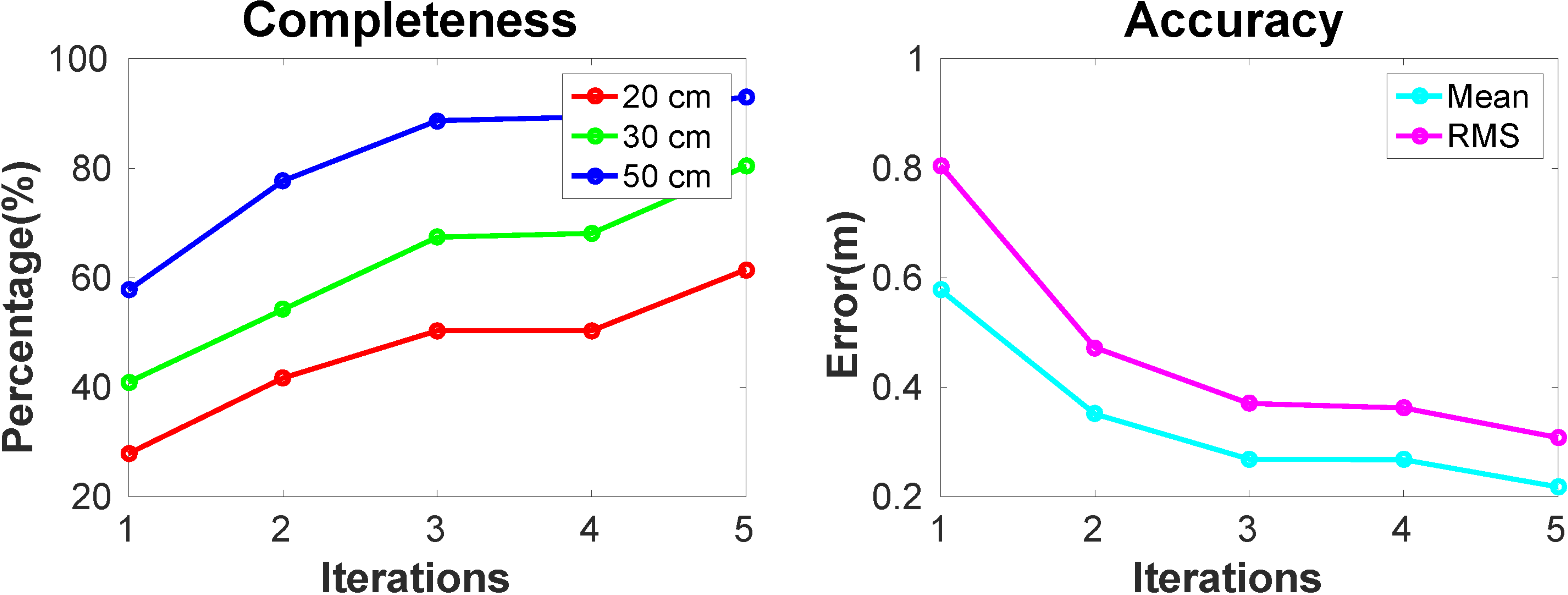}
\caption{The \emph{Asian Building} example. Model completeness and accuracy over iterations.}
\label{fig:gate_gt}
\vspace{-0.2in}
\end{figure}

\paragraph*{\emph{Outdoor Experiments}} For outdoor scenes, we compare our results with a model generated from exhaustive image capture. 
Our system captures 101 views of the model in 5 iterations. We manually captured 328 views to generate the ground truth model. \Fig{fig:gate_gt} shows that the model completeness and accuracy improves over the iterations.

\section{Conclusion}
This paper presents a method to automate the image-capturing process in large scale image-based modeling.
Technically, it contributes a fast MVS method and an efficient NBV algorithm.
The MVS problem is solved by an iterative linear method, which makes online model reconstruction and coverage assessment possible. The NBV algorithm benefits from our customized coverage evaluation method, and adopts an efficient search strategy. Experiments on real examples have demonstrated the effectiveness of our system.

\bibliographystyle{IEEEtran}
\bibliography{IEEEabrv,references}

\end{document}